\documentclass[runningheads]{llncs}
\usepackage[T1]{fontenc}
\usepackage{graphicx}

\usepackage{cite}
\usepackage{amsmath,amssymb,amsfonts}
\usepackage{algorithmic}
\usepackage{graphicx}
\usepackage{textcomp}
\usepackage{xcolor}
\usepackage{subcaption}
\usepackage{tikz}
\usetikzlibrary{arrows.meta, positioning}
\usepackage[table]{xcolor}

\begin{document}
%
\title{Measuring What Matters: A Causal Reliance Score for Histopathology Image Segmentation}
\titlerunning{A Causal Reliance Score for Histopathology Segmentation}
%
\author{Samuel Thomas\inst{1} \and
Daniel Tweneboah Anyimadu\inst{1} \and
Ahmed Karam~Eldaly\inst{1} \and
Somaia Elsheikh\inst{2} \and
Mohammed M. Abdelsamea}
\authorrunning{S. Thomas et al.}
%
\institute{Department of Computer Science, University of Exeter, Exeter, UK \\
\email{\{scct201, da536, a.karam-eldaly, m.abdelsamea\}@exeter.ac.uk}\\
\and
Nottingham University Hospitals NHS Trust, 
\email{\{somaia.elsheikh\}@nhs.net}}
\maketitle              
%

\begin{abstract}
Reliable histopathology segmentation requires models that rely on pathology-relevant structures rather than spurious background correlations. In this work, we argue that improving architectures alone is insufficient if models continue to rely on spurious correlations rather than pathology-relevant features. We introduce a causality-inspired evaluation framework based on controlled interventions that explicitly probes model sensitivity to foreground tissue structure and background/domain variation. Building on this, we propose the Causal Reliance Score (CRS), a metric that quantifies the relative sensitivity of model predictions to foreground perturbation versus background/domain variation under controlled interventions. To support this analysis, we design augmentation strategies that act as implicit counterfactual interventions during training, encouraging invariance to nuisance factors while preserving morphology. We demonstrate how these interventions reveal behavioural differences between models that are not captured by standard metrics such as Dice or IoU. While we instantiate our framework using a self-supervised transformer (CAUSE-ViT), the primary contribution of this work is not a new architecture, but a principled intervention-based methodology and evaluation metric for auditing model behaviour in medical image segmentation. Our results suggest that evaluating model behaviour through counterfactual interventions is essential for developing reliable and clinically robust medical image segmentation systems.

\end{abstract}

\section{Introduction}
Deep learning (DL) has become central to medical image analysis, enabling significant advancements in automated annotation, inference efficiency, and quantitative evaluation through standardised metrics \cite{kar2021review}. However, the widespread deployment of these tools in clinical settings remains constrained by challenges such as small and heterogeneous datasets, privacy concerns, and the need for expert annotation, all of which limit model generalisation \cite{lopez2019medical, tajbakhsh2020embracing}. Additionally, variations in imaging protocols, staining procedures, and acquisition techniques across institutions introduce biases, leading models to overfit spurious correlations rather than learning clinically meaningful features \cite{lopez2019medical}.

These challenges are particularly pronounced in histopathology segmentation, where the complexity of tissue morphology, extreme class imbalance, and the need for pixel-level annotations make segmentation tasks difficult \cite{lopez2019medical}. Existing DL models, such as fully convolutional networks (FCNs) \cite{long2015fully} and U-Net variants \cite{ronneberger2015u}, often struggle to capture the global spatial patterns essential for accurate segmentation. Instead, these models tend to exploit local texture cues or biased receptive fields, which are heavily influenced by dataset artifacts, undermining their ability to generalise across diverse populations. Vision Transformers (ViTs) \cite{vaswani2017attention}, originally developed for natural language processing, have recently shown strong performance in computer vision by capturing long-range dependencies and global context, which are valuable for dense prediction tasks like histopathology segmentation. Unlike convolutions, ViTs use self-attention mechanisms that model spatial relationships across the entire image, producing richer feature representations and overcoming some of the limitations of traditional convolutional neural networks (CNNs)  \cite{dosovitskiy2020image, xiao2023transformers}.

In light of these challenges, self-supervised learning (SSL) has emerged as a promising paradigm to mitigate data scarcity by leveraging unlabelled data to improve representation learning and reduce reliance on extensive annotation. When combined with ViTs, SSL enables the modelling of long-range dependencies and global context, which are especially important for histopathology segmentation. However, despite its potential, existing SSL methods face challenges in ensuring that models learn robust, pathology-aligned representations. Without careful pretraining objectives or augmentation strategies, there remains a risk that models may reinforce shortcut learning, focusing on background cues rather than pathology-relevant features. Generative SSL, such as Masked Autoencoders (MAE), and contrastive SSL approaches help improve model performance by promoting spatially coherent and context-aware representations, yet these techniques often fail to fully account for pathology-aligned learning.

In this work, we take a different perspective: rather than proposing a new architecture as the primary contribution, we focus on understanding and evaluating model behaviour through an intervention-based, causality-inspired lens. We introduce counterfactual interventions, implemented via targeted augmentations, to explicitly test whether models rely on pathology-relevant structures or spurious correlations. This shifts the focus from improving predictive performance alone to auditing why models make their predictions. The main contributions of this work are:
\begin{itemize}
\item We introduce an intervention-based evaluation framework for histopathology segmentation, where targeted  perturbations are used to probe model reliance on foreground tissue structure versus background context.
\item We propose the Causal Reliance Score (CRS), a novel evaluation metric that quantifies the relative sensitivity of models to foreground tissue perturbation and nuisance background/domain variation, complementing standard segmentation metrics.
\item We provide an experimental instantiation using a self-supervised transformer (CAUSE-ViT), demonstrating how supervised, generative and contrastive training strategies differ under controlled behavioural probes.
\end{itemize}

\noindent Thus, unlike prior work that focuses on improving segmentation accuracy through architectural design or training objectives, this work introduces a fundamentally different perspective: evaluating what models rely on when making predictions. To our knowledge, this is among the early efforts to frame controlled intervention-based behavioural probing as a systematic evaluation protocol for medical image segmentation.

\section{Related Works}
Recent advances in medical image segmentation have increasingly leveraged Vision Transformers to address the limitations of convolutional neural networks in capturing long-range spatial dependencies \cite{dosovitskiy2020image, xiao2023transformers, thisanke2023semantic}. Architectures such as SETR \cite{zheng2021rethinking} and Segmenter \cite{strudel2021segmenter} reformulate segmentation as a sequence-to-sequence prediction task, enabling global context modelling through self-attention. These approaches have demonstrated strong performance in dense prediction tasks, including histopathology segmentation, where global tissue structure is crucial. However, their effectiveness often depends on large-scale training data \cite{berroukham2023vision} and carefully designed decoding strategies to recover fine-grained spatial detail \cite{xiao2023transformers}.

To mitigate data scarcity, self-supervised learning (SSL) has emerged as a key paradigm for representation learning in medical imaging. Contrastive methods, such as LETCP \cite{chen2022letcp}, learn invariant embeddings by maximising agreement between augmented views of the same image while separating unrelated samples. Generative approaches, including Masked Autoencoders (MAE) \cite{he2022masked}, reconstruct masked image regions to capture global structural information. These strategies have been shown to improve downstream segmentation performance and robustness, particularly under limited annotation settings \cite{liu2021self}. 

Despite these advances, existing transformer-based and SSL-driven segmentation methods primarily optimise for predictive performance and do not explicitly assess whether learned representations align with pathology-relevant structures. In histopathology, this limitation is critical, as models may achieve high accuracy while relying on spurious correlations introduced by staining variations, acquisition protocols, or dataset-specific artefacts. Recent work in domain generalisation and shortcut learning highlights the need for evaluating model behaviour beyond standard metrics, emphasising robustness and invariance to nuisance factors \cite{abdelsamea2026causal}. In spite of the growing interest in robustness and domain generalisation, there remains a paucity of systematic frameworks for explicitly probing model reliance in medical image segmentation. Existing approaches evaluate performance under shift, but do not isolate the causal factors driving model predictions.

In contrast to prior work, which primarily focuses on architectural improvements or performance gains, our approach emphasises evaluation and analysis. We introduce a counterfactual, intervention-based framework that enables direct probing of model behaviour under controlled perturbations. This positions our work as complementary to existing segmentation methods, providing tools to assess how and why models make predictions, rather than proposing yet another model to improve benchmark performance.

\section{Counterfactual Intervention Framework}
This section introduces the core contribution of the paper: a causality-inspired, intervention-based framework for analysing model behaviour. The proposed architecture (CAUSE-ViT) serves as a testbed to study how different training strategies interact with these controlled perturbations.

\subsection{Causality-Inspired Reliance Paradigm}
\begin{figure}[h!]
    \centering
    \includegraphics[width=1.\linewidth]{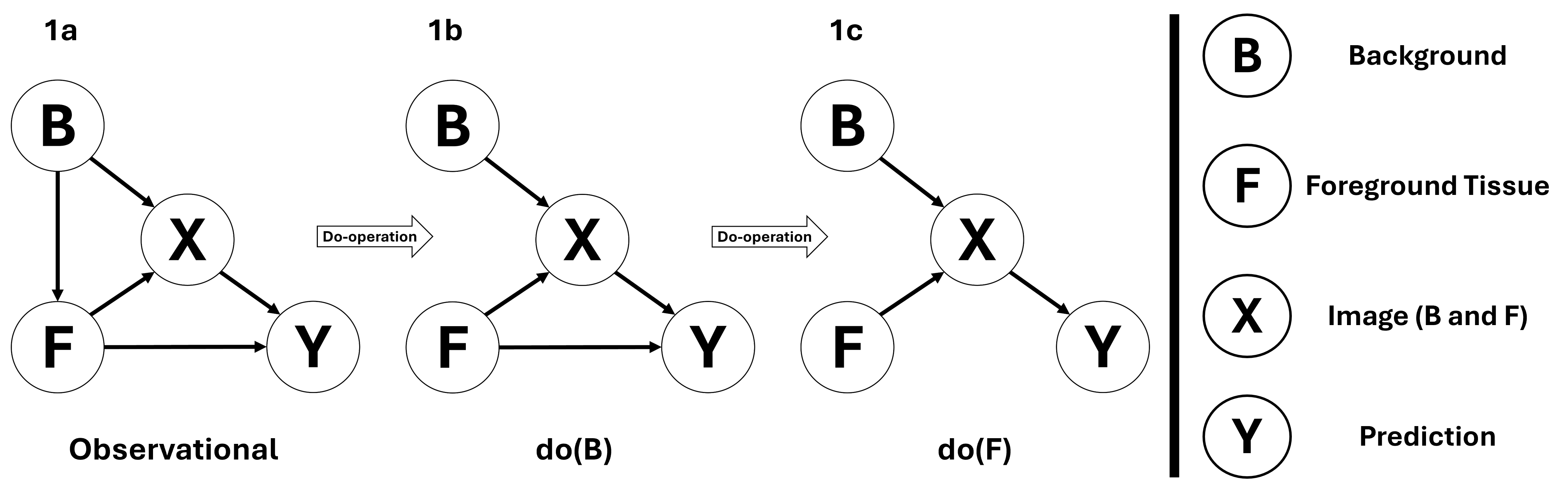}
    \caption{ Conceptual abstraction and intervention protocol. Foreground tissue structure ($F$) and background context ($B$) jointly shape the observed image ($X$), which is mapped to the segmentation prediction ($Y$), i.e., $F, B \rightarrow X$ and $X \rightarrow Y$. Panel (1a) shows the observational setting, whereas Panels (1b) and (1c) illustrate controlled perturbations of background and foreground structure. Resulting changes in prediction are used as an empirical probe of model sensitivity to pathology-relevant tissue structure and background/domain variation.}
    \label{fig:base_causal_graph}
\end{figure}

CAUSE-ViT is designed to address bias and spurious correlations in histopathology segmentation, where models may overfit to irrelevant background cues. To this end, we introduce a causality-inspired, intervention-based evaluation framework that distinguishes pathology-relevant foreground structure from nuisance background variation. We adopt a conceptual abstraction (Figure \ref{fig:base_causal_graph}) in which foreground tissue structure ($F$) and background context ($B$) jointly shape the observed histopathology image ($X$), which is then mapped to a segmentation prediction ($Y$). 
This abstraction serves as a motivating lens for designing controlled perturbations that probe whether model predictions are more sensitive to tissue-relevant structure or to spurious contextual signals.

In practice, these probes are implemented through image-space perturbations, such as local blurring of foreground regions or modification of background appearance. We emphasise that this framework is causality-inspired rather than a formal causal inference model: the perturbations are used as controlled behavioural probes, not as interventions derived from an identified structural causal model. The resulting changes in prediction, measured through intervention-induced Dice contrasts, therefore provide an empirical proxy for the model’s relative sensitivity to foreground perturbation and background/domain variation.

\subsection{CAUSE-ViT Architecture}
\begin{figure}[ht!]
    \centering
    \begin{subfigure}[b]{1\linewidth}
        \centering
        \includegraphics[width=\linewidth]{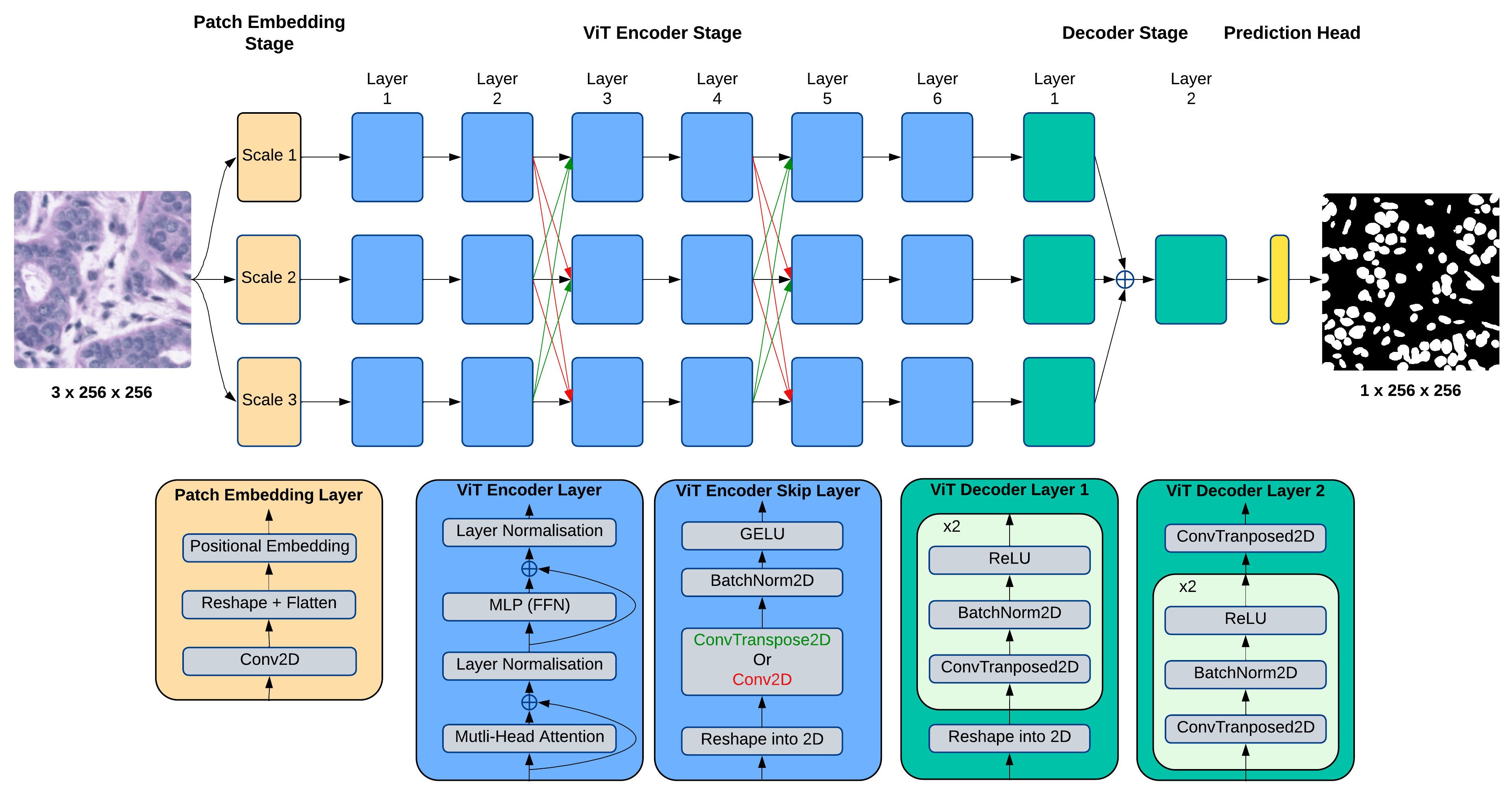}
        \caption{CAUSE-ViT architecture with multi-scale patch embedding, cross-scale ViT encoding, and decoder-based segmentation prediction.}
        \label{fig:model_design}
    \end{subfigure}
    \hfill
    
    \begin{subfigure}[b]{1\linewidth}
        \centering
        \includegraphics[width=\linewidth]{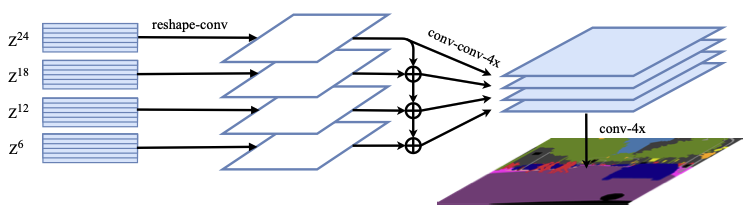}
        \caption{SETR-MLA-inspired decoder head for segmentation \cite{zheng2021rethinking}.}
        \label{fig:setr_mla_decoder}
    \end{subfigure}    
    \caption{High-level design overview of CAUSE-ViT.
    }
    \label{fig:combined_model_ssl}
\end{figure}

We introduce CAUSE-ViT as a representative transformer-based segmentation model used to instantiate and validate the proposed evaluation framework. The architectural design is intentionally aligned with existing multi-scale ViT paradigms, allowing us to isolate the effect of training strategies and interventions. At the core of CAUSE-ViT lies a multi-scale self-supervised Vision Transformer that follows an encoder–decoder design (Figure \ref{fig:combined_model_ssl}). The key novelty of CAUSE-ViT is a multi-scale encoder composed of independent scale-specific ViT streams with periodic cross-scale feature fusion, enabling structured interaction between representations at different resolutions. Specifically, (1) the encoder consists of multiple parallel ViT streams, each operating at a distinct resolution to capture complementary morphological patterns. Unlike conventional multi-scale designs that rely on shared weights or late fusion, each stream in CAUSE-ViT is independently initialised, encouraging scale-specific feature learning. (2) To promote interaction across scales, we introduce periodic cross-scale feature fusion, where representations are exchanged between streams every two layers. 
This allows the model to jointly capture fine-grained cellular details and broader tissue context, improving representation consistency across resolutions. Building on this encoder, CAUSE-ViT employs a high-resolution ViT (HRViT)-style backbone to preserve local detail (Figure \ref{fig:model_design}), while integrating a SETR-MLA-inspired decoder (Figure \ref{fig:setr_mla_decoder}) to aggregate multi-scale features into coherent global representations. Skip connections, implemented via transposed or standard convolutions, ensure efficient information flow and preserve spatial detail during upsampling. The decoder upsamples encoder features through learnable transposed convolutions, followed by ReLU activations and normalisation, avoiding information loss associated with fixed interpolation methods, as in SegFormer \cite{xie2021segformer}. Finally, a $2$D convolution projects the output into per-pixel class probabilities, producing a segmentation mask of size $(Num_{Classes}, Out_w, Out_h)$ from an input image of size $(RGB, In_w, In_h)$. Overall, this design encourages representations that are less dependent on scale-specific artefacts and more consistent across resolutions, supporting robust segmentation under domain variability.

\subsection{Self-supervised Pretraining}
\begin{figure}[ht!]
    \centering    

    \begin{subfigure}[b]{0.53\linewidth}
        \centering
        \includegraphics[width=\linewidth]{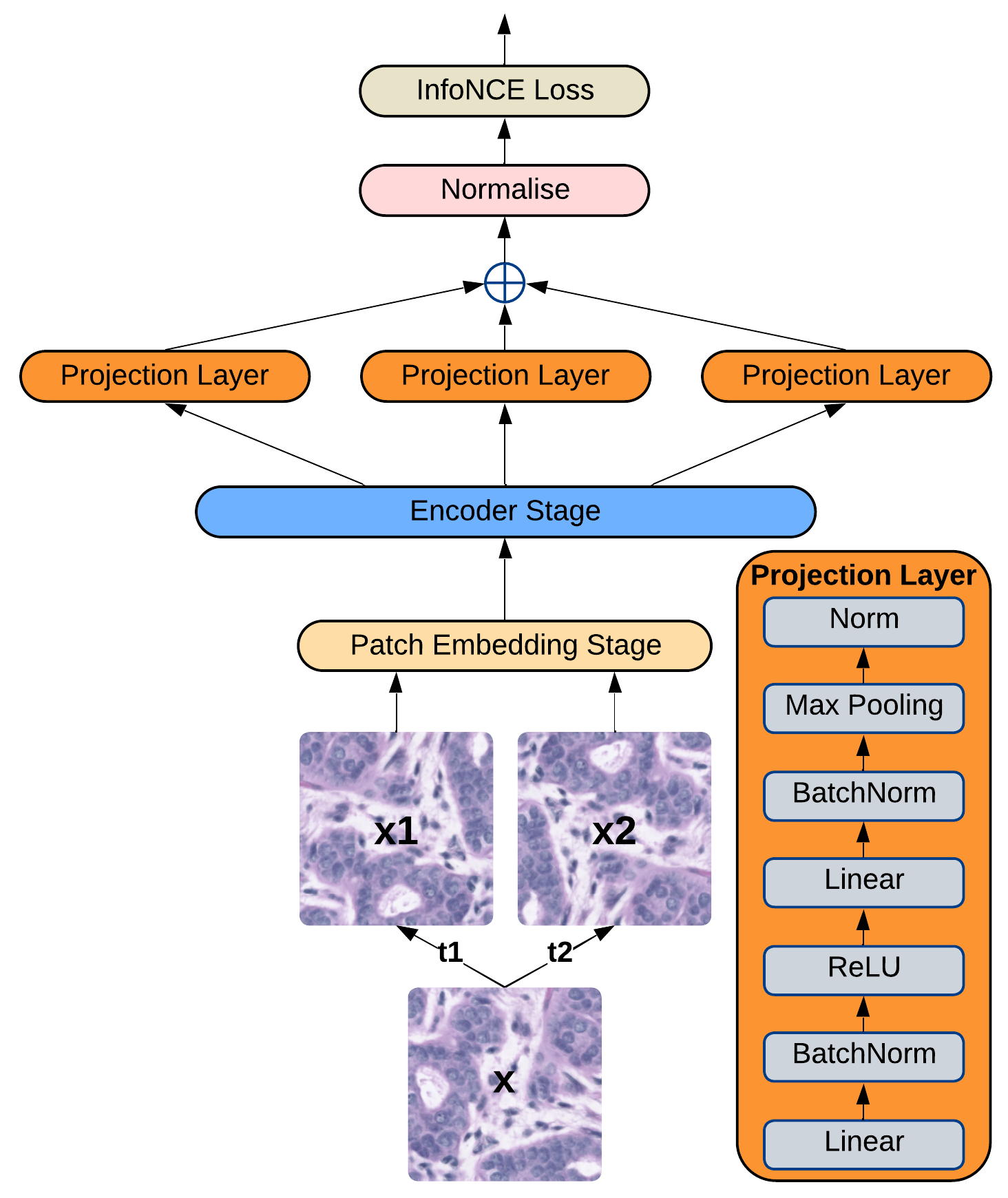}
        \caption{Contrastive SSL Wrapper Approach.}
        \label{fig:contrastive_ssl_diagram}
    \end{subfigure}
    \hfill
    
    \begin{subfigure}[b]{0.35\linewidth}
        \centering
        \includegraphics[width=\linewidth]{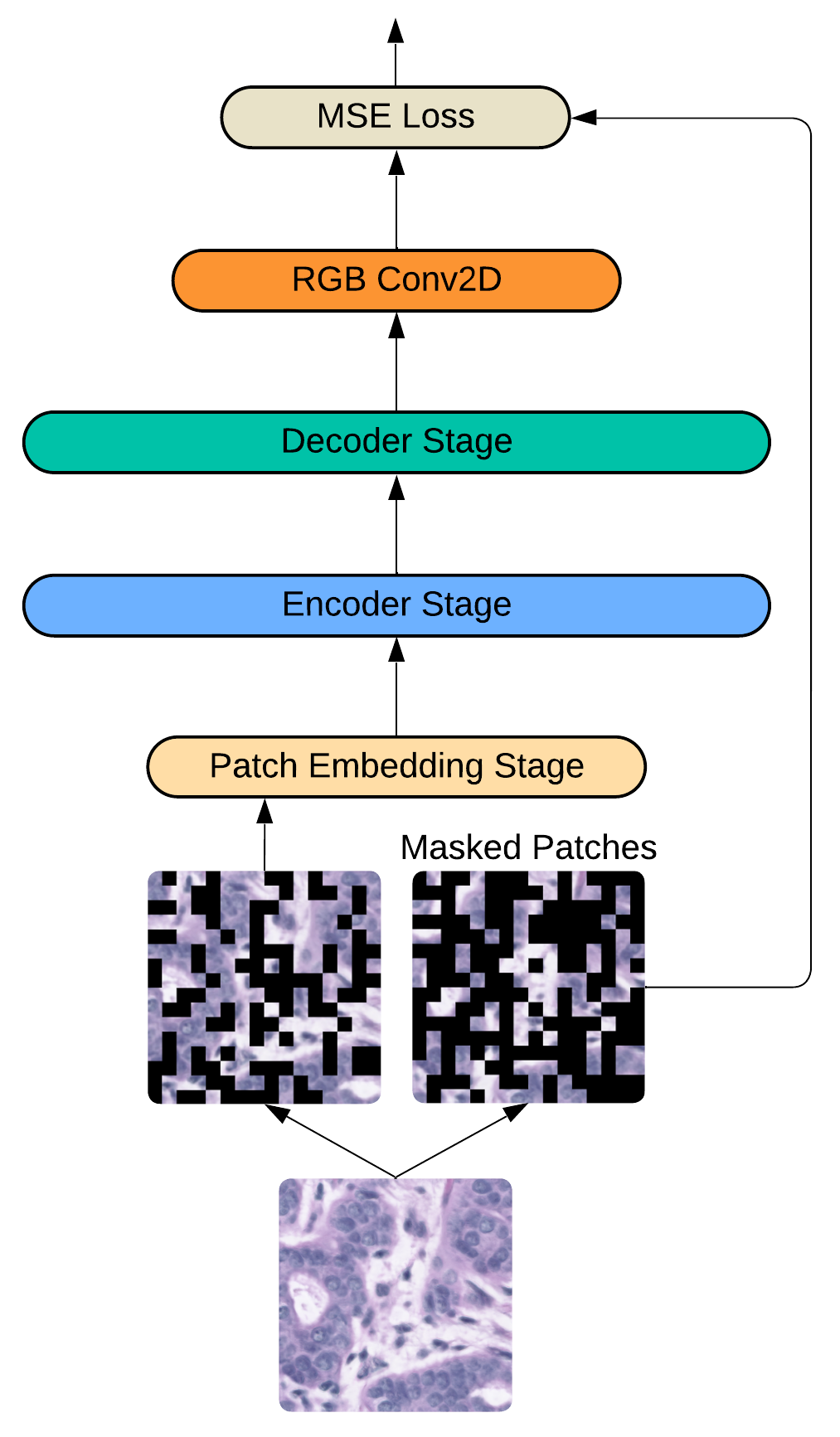}
        \caption{Generative SSL Wrapper Approach.}
        \label{fig:generative_ssl_diagram}
    \end{subfigure}

    \caption{High-level design overview of SSL wrapper models.
    }
    \label{fig:combined_ssl_wrapper}
\end{figure}

To mitigate the dependence on extensive annotations, CAUSE-ViT is pretrained using self-supervised learning. Two complementary SSL strategies are employed to enhance the model's ability to learn robust features from unlabelled data. (1) Contrastive SSL (Figure \ref{fig:contrastive_ssl_diagram}), inspired by LETCP \cite{chen2022letcp}, learns discriminative representations by generating positive and negative pairs through controlled transformations of the same image. These transformations include spatial augmentations (e.g., $90^\circ$ rotations) and non-spatial (appearance-based) augmentations (e.g., Gaussian blur, colour jittering, and illumination adjustment), which simulate variations in morphology, staining, and acquisition conditions commonly observed in histopathology \cite{veta2014breast}. A projection head (linear layer–Batch Normalisation (BN)–ReLU activation) followed by max-pooling encourages the encoding of salient morphological patterns, which are optimised using the InfoNCE loss. (2) Generative SSL (Figure \ref{fig:generative_ssl_diagram}), based on Masked Autoencoders (MAE) \cite{he2022masked}, learns contextual representations by masking patches of the input image and reconstructing them from the visible regions. Fixed patch sizes maintain compatibility with skip connections, while omitting positional embeddings encourages the encoding of structural patterns rather than absolute spatial positions. Masking larger patches exposes both full and partial cellular structures, improving robustness to missing or noisy regions. Prior to both contrastive and generative SSL, the aforementioned augmentations are applied as input-level transformations, either globally or selectively to background regions, preserving morphological integrity while exposing the model to realistic variations without introducing artefacts. Together, these SSL strategies enable CAUSE-ViT to learn representations that are less sensitive to nuisance variations and more aligned with pathology, even under limited supervision.

\subsection{Causality-Inspired Interventional Evaluation}
We now formalise the proposed counterfactual evaluation framework. Unlike standard evaluation protocols that measure predictive accuracy, our approach explicitly measures how predictions change under controlled interventions on the input. This enables direct assessment of whether models rely on pathology-relevant structures or spurious correlations. While augmentations are commonly used for regularisation during training, we reinterpret them here as explicit interventions at evaluation time; rather than improving performance,  these interventions are used to probe model behaviour under controlled distribution shifts. Accordingly, we define a conceptual structural abstraction with foreground tissue structure ($F$), background context ($B$), observed image ($X$), and segmentation output ($Y$), with causal relations $F,B \rightarrow X$ and $X \rightarrow Y$.\\

\noindent To disentangle sensitivity to nuisance background variation from sensitivity to pathology-relevant foreground structure, we define the intervention protocol using a matched reference condition. First, we construct a background-normalised image, $X^{ref}$, in which background/domain appearance is adjusted while foreground tissue morphology is preserved. We then generate a foreground-perturbed image, $X^{fg}$, by applying local blurring to the foreground regions of $X^{ref}$ while keeping the background treatment unchanged.
The foreground intervention effect is defined as
\begin{equation}
\mathrm{CE}_{F} = \mathbb{E}\big[\mathrm{Dice}(Y(X^{ref}),M) - \mathrm{Dice}(Y(X^{fg}),M)\big],
\end{equation}
where $M$ denotes the ground-truth mask. This measures the performance decrease attributable to degradation of foreground tissue structure under matched background conditions.
The background sensitivity term is defined relative to the standard out-of-distribution (OOD) image, $X^{ood}$, as
\begin{equation}
\mathrm{CE}_{B} = \mathbb{E}\big[\mathrm{Dice}(Y(X^{ref}),M) - \mathrm{Dice}(Y(X^{ood}),M)\big],
\end{equation}
which quantifies the performance difference associated with nuisance background/domain variation.
The Causal Reliance Score is then given by
\begin{equation}
\mathrm{CRS} = \frac{\mathrm{CE}_{F}}{\mathrm{CE}_{B} + \epsilon}.
\end{equation}
Under this formulation, CRS is interpreted as a relative sensitivity measure with respect to the chosen perturbations, separating the contribution of background/domain mismatch from the incremental effect of foreground degradation.

\section{Experiments and Results}
The goal of the experiments is to demonstrate the effectiveness of the proposed counterfactual intervention framework and CRS in revealing model behaviour.
\subsection{Datasets}
The proposed CAUSE-ViT framework is evaluated under both in-distribution (InD) and out-of-distribution (OOD) histopathology settings to assess performance under matched and shifted data conditions. For InD training and validation, we use SNOW (Synthetic Nuclei and annotation Wizard) \cite{ding2023large}, a large-scale synthetic breast histology dataset of $512\times512$ image tiles with paired nuclei segmentation masks, which were resized to $256\times256$ and randomly augmented to support stable model development. For OOD evaluation, we use CryoNuSeg \cite{mahbod2021cryonuseg}, a fully manually annotated frozen-section H\&E dataset that introduces realistic domain shifts, including organ diversity, staining variability, and frozen-section artefacts. This dataset pairing is intentional: SNOW enables reliable optimisation under controlled conditions, while CryoNuSeg provides a stringent test of whether the model generalises through pathology-relevant features rather than dataset-specific background or staining cues. In total, SNOW contains 18,213 images, which are split into 80\% for training and 20\% for validation.

\subsection{Evaluation Criteria, Interventional Robustness, and Causal Reliance}
Segmentation performance is evaluated using binary cross-entropy (BCE) loss, Dice score, and mean Intersection over Union (mIoU) across both InD and OOD settings. To assess robustness under causal perturbations, we further evaluate the model under Causal Intervention (1b: Raw OOD input), Causal Intervention (1b: Background-normalised reference), and Causal Intervention (1c: Foreground-blurred condition). Beyond accuracy, we analyse model sensitivity to foreground tissue structure and background/domain variation using the proposed CRS, computed post-hoc on CryoNuSeg from Dice-based interventional contrasts. Specifically, evaluation on CryoNuSeg is performed under three conditions: the raw OOD input, a background-normalised reference condition that preserves foreground tissue structure, and a matched foreground-blurred condition obtained by applying local blurring to the cell regions of the same reference image. In line with the formulation above, the reference condition is used to separate the contribution of nuisance background/domain variation from the incremental effect of foreground degradation. Accordingly, $\mathrm{CE}_{B}$ captures the performance difference between the reference condition and the raw OOD input, whereas $\mathrm{CE}_{F}$ captures the additional performance decrease induced by foreground blurring relative to that same reference. CRS is then computed from these two interventional contrasts. Thus, unlike Dice or mean IoU, CRS cannot be computed from standard predictions alone; it requires evaluation under a controlled perturbation protocol and is therefore intended as an intervention-based companion metric for behavioural auditing on held-out test data.

\subsection{Implementation Details}
All experiments were implemented in Python and conducted on Google Colab using NVIDIA A100 and L4 GPUs. CAUSE-ViT, comprising 74M parameters with a multi-scale patch embedding module, 8 attention heads, and a dropout rate of 0.2, was trained under three experimental settings: supervised learning, generative SSL, and contrastive SSL. The supervised configuration was trained for 25 epochs with a batch size of 32, while both generative SSL and contrastive SSL were trained for 50 epochs, using batch sizes of 32 and 45, respectively.

\subsection{Quantitative Results}
Table \ref{tab:all_two_summary} summarises the segmentation performance of the CAUSE-ViT variants using BCE loss, Dice, and mean IoU on the SNOW InD dataset. The supervised baseline learns directly from labelled data, which can lead to reliance on dataset-specific appearance cues, while contrastive SSL leverages paired augmentations to enforce representation consistency but may remain sensitive to superficial visual correlations. In contrast, generative SSL, based on masked reconstruction, encourages the model to recover missing structural information, thereby promoting the learning of more intrinsic and pathology-relevant features. As a result, CAUSE-ViT with generative SSL consistently performs best on the SNOW InD dataset, indicating stronger segmentation quality and improved generalisation to unseen hold-out validation data. Table \ref{tab:all_intervention_summary} further evaluates the same models on CryoNuSeg under causal intervention conditions. Both the background-normalised and foreground-blurred conditions produce substantially higher Dice scores than the raw OOD baseline, indicating that the raw OOD setting is dominated by nuisance background/domain mismatch and that the principal performance recovery is associated with background normalisation. Foreground sensitivity is therefore assessed relative to the matched reference condition, under which foreground blurring induces only a modest additional Dice decrease. Consistent with this pattern, CRS values are low across all models, indicating that the dominant performance change arises from background/domain variation rather than from the applied foreground perturbation. The Generative SSL variant attains the lowest CRS, reflecting the smallest incremental sensitivity to foreground blurring relative to the much larger background/domain effect. Together with its consistently strongest Dice scores across conditions, this suggests that Generative SSL is the most stable model under the present intervention design. However, because the current foreground perturbation produces only a limited additional decrease, these results should be interpreted as evidence of relative robustness under intervention rather than as a strong demonstration of dominant foreground reliance.

\begin{table}[h!]
    \small
    \centering
    \caption{Segmentation performance of CAUSE-ViT variants on the SNOW (InD) dataset. Best results in bold.}
    \resizebox{\linewidth}{!}{ 
    \begin{tabular}{|l|c|c|c|}
        \hline
        \textbf{Metrics} & \textbf{Supervised} & \textbf{CAUSE-ViT Generative SSL} & \textbf{CAUSE-ViT Contrastive SSL} \\ \hline\hline

        \multicolumn{4}{|c|}{\textbf{SNOW Dataset Validation}} \\ \hline
        BCE Loss (↓) & 0.1354 & \textbf{0.1303} & 0.1360 \\
        Dice (↑)     & 0.8661 & \textbf{0.8752} & 0.8650 \\
        Mean IoU (↑) & 0.7887 & \textbf{0.8046} & 0.7873 \\ \hline
        
    \end{tabular}
    }
    \label{tab:all_two_summary}
\end{table}

\begin{table}[h!]
    \small
    \centering
    \caption{ Segmentation performance of CAUSE-ViT variants on CryoNuSeg under three evaluation conditions: raw OOD input, background-normalised reference, and matched foreground blur. This setup separates background/domain effects from foreground sensitivity. Best results are shown in bold.}
    \resizebox{\linewidth}{!}{ 
    \begin{tabular}{|l|c|c|c|}
        \hline
        \textbf{Metrics} & \textbf{Supervised} & \textbf{CAUSE-ViT Generative SSL} & \textbf{CAUSE-ViT Contrastive SSL} \\ \hline\hline

       \multicolumn{4}{|c|}{\textbf{ Causal Intervention 1b (Raw OOD Evaluation)}} \\ \hline
        BCE Loss (↓) & 1.6284 & \textbf{0.8520} & 1.6921 \\
        Dice (↑)     & 0.0228 & \textbf{0.2492} & 0.0008 \\
        Mean IoU (↑) & 0.0116 & \textbf{0.1423} & 0.0004 \\ \hline

        \multicolumn{4}{|c|}{\textbf{ Causal Intervention 1b (Background-Normalised Reference)}} \\ \hline
        BCE Loss (↓) & 0.1553 & \textbf{0.0879} & 0.1514 \\
        Dice (↑)     & 0.8981 & \textbf{0.9370} & 0.8930 \\
        Mean IoU (↑) & 0.8150 & \textbf{0.8815} & 0.8067\\ \hline

        \multicolumn{4}{|c|}{\textbf{Causal Intervention 1c (Foreground-Blurred Condition)}} \\ \hline
        BCE Loss (↓) & 0.1666 & \textbf{0.0900} & 0.1587 \\
        Dice (↑)     & 0.8918 & \textbf{0.9352} & 0.8888  \\
        Mean IoU (↑) & 0.8047 & \textbf{0.8783} & 0.7998 \\ \hline\hline
        
        \multicolumn{4}{|c|}{\textbf{ Causal Reliance Score (CRS)}} \\ \hline
        CRS (↓) & 0.0072 & \textbf{0.0026} & 0.0047 \\ \hline
    \end{tabular}
    }
    \label{tab:all_intervention_summary}
\end{table}

Taken together, these results show that the proposed framework provides a more complete assessment of model behaviour under distribution shift. An important distinction is that Dice and CRS do not rank models according to the same criterion. Dice measures segmentation accuracy under a specific evaluation condition, whereas CRS summarises the relative effect of foreground perturbation compared with background/domain variation. As a result, a model may achieve slightly higher Dice yet also exhibit a higher CRS, or vice versa, without contradiction. For example, although the supervised model attains slightly higher Dice than the contrastive model under the background-normalised and foreground-blurred conditions, its CRS is also higher because the incremental foreground-blur effect is larger relative to the background/domain effect. Thus, CRS should not be interpreted as a replacement for Dice, but as a complementary intervention-based metric that provides behavioural context for the observed segmentation performance. This highlights the benefit of complementing standard accuracy metrics with intervention-based analysis when the aim is to understand how model performance changes under controlled perturbations.

\subsection{Qualitative Results}

\begin{figure*}[htbp]
    \centering
    \includegraphics[width=1.\linewidth]{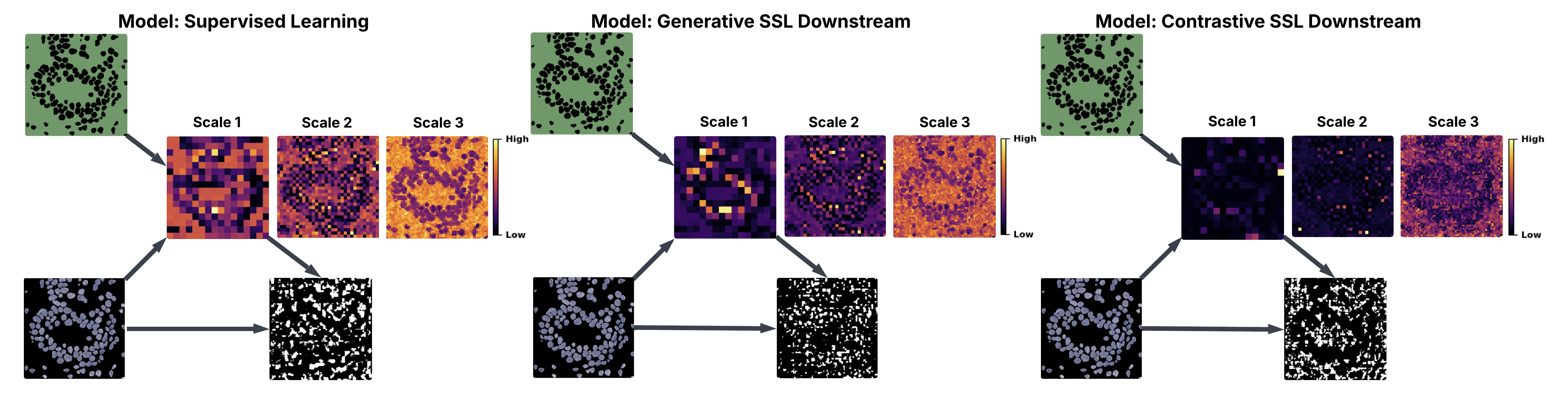}
    \caption{Causal Intervention 1b: raw OOD input (background highlighted in amber). Columns show supervised, generative SSL, and contrastive SSL models, together with the input image, multi-scale responses (Scales 1–3), reference foreground structure, and predicted mask. The generative SSL model shows comparatively stronger foreground localisation under OOD shift.}
    \label{fig:causal_intervention_bg_no_aug}
\end{figure*}

\begin{figure*}[h!]
    \centering
    \includegraphics[width=1.\linewidth]{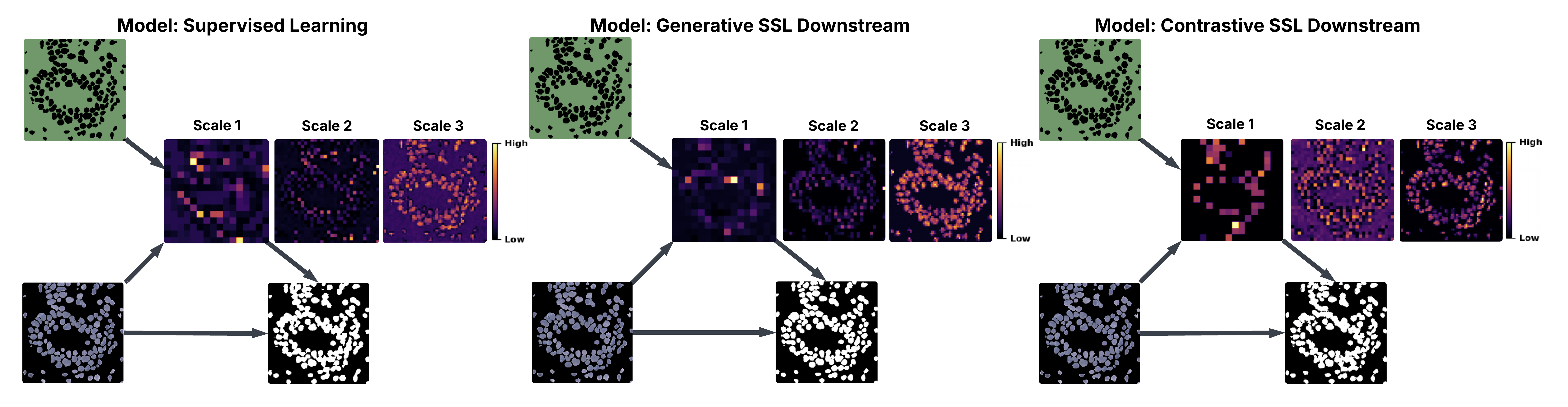}
    \caption{Causal Intervention 1b: background-normalised reference condition (background perturbed while foreground tissue structure is preserved; highlighted in amber). Columns show supervised, generative SSL, and contrastive SSL models, together with the input image, multi-scale responses (Scales 1–3), predicted mask, and reference foreground mask. The generative SSL model shows the most stable response under background/domain correction.}
    \label{fig:causal_intervention_bg_aug}
\end{figure*}

\begin{figure*}[h!]
    \centering
    \includegraphics[width=1.\linewidth]{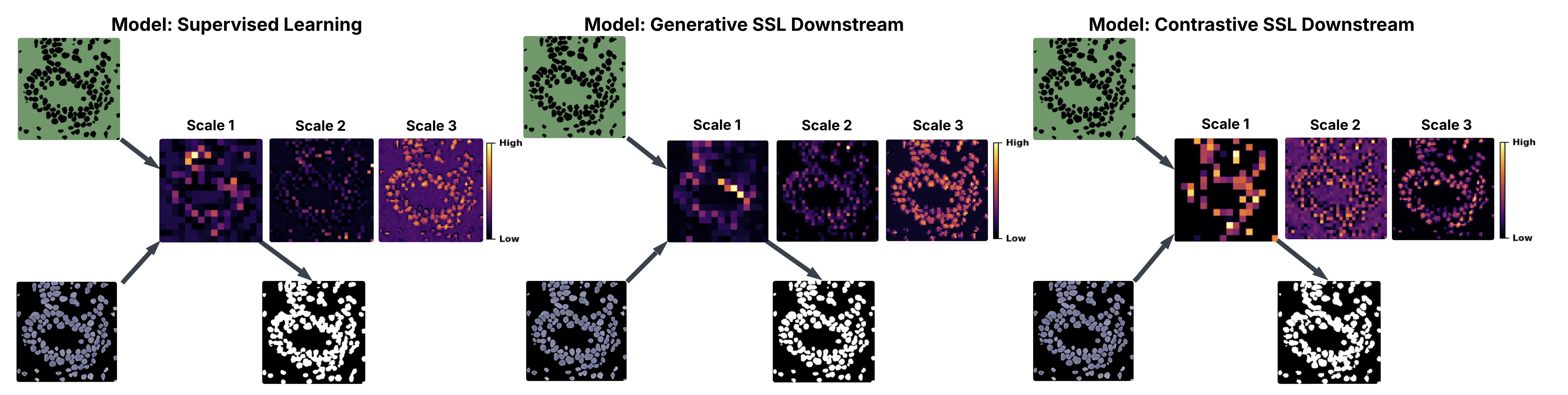}
    \caption{Causal Intervention 1c: foreground-blurred condition (background-normalised reference with additional Gaussian blurring of foreground tissue structure). Columns show supervised, generative SSL, and contrastive SSL models, together with the input image, multi-scale responses (Scales 1–3), predicted mask, and reference foreground mask. The generative SSL model remains comparatively stable under the applied foreground perturbation.}
    \label{fig:causal_intervention_bg_fg_aug}
\end{figure*}

Figures (\ref{fig:causal_intervention_bg_no_aug}, \ref{fig:causal_intervention_bg_aug}, and \ref{fig:causal_intervention_bg_fg_aug}),  illustrate the causal intervention experiments on the CryoNuSeg OOD dataset. Without augmentation (Figure \ref{fig:causal_intervention_bg_no_aug}), all models become highly dependent on background context, failing to detect pathology-relevant foreground regions. Generative SSL, however, shows reduced background bias and comparatively higher scores. After applying augmentations (Figure \ref{fig:causal_intervention_bg_aug}), models shift attention to tissue-relevant structure (pathology-relevant cells), with SSL, especially the generative variant, achieving the best performance. When additional Gaussian blurring of pathology-relevant foreground cells is introduced, performance remains relatively close to that of the background-normalised reference condition. This suggests that the applied perturbation does not fully disrupt the cues used by the model and may indicate reliance on coarser morphological structure in addition to fine local texture. However, this stability should be interpreted cautiously, because the dominant improvement relative to the raw OOD baseline is driven by background/domain correction rather than by the foreground intervention itself. Overall, SSL methods, particularly the generative variant, show greater stability under the intervention protocol and stronger robustness to the observed OOD background/domain variation.

\section{Conclusion}
This work introduces a counterfactual intervention framework for analysing model behaviour in histopathology segmentation. Rather than focusing on architectural novelty, we emphasise understanding how model behaviour changes under controlled interventions. Through the proposed CRS and intervention-based analysis, we show that standard metrics such as Dice and IoU are insufficient to explain performance changes under distribution shift. In particular, robust interpretation requires separating nuisance background/domain effects from foreground perturbation effects. Our results demonstrate that models with similar performance can exhibit fundamentally different behaviours under perturbation. We further show that augmentation strategies can be interpreted as implicit counterfactual interventions, providing a bridge between training and evaluation. While demonstrated using a transformer-based model, the proposed framework is model-agnostic and can be applied broadly to audit reliability and robustness in medical image segmentation systems. The proposed framework relies on approximated interventions implemented via image transformations, which may not perfectly correspond to true causal manipulations. Additionally, the CRS is a relative metric and depends on the choice of perturbations. Future work will explore more principled causal formulations and extend the framework to other imaging modalities and tasks.

\bibliographystyle{splncs04}
\bibliography{references}

@article{abdelsamea2026causal,
  title={Causal Transfer in Medical Image Analysis},
  author={Abdelsamea, Mohammed M and Anyimadu, Daniel Tweneboah and Selim, Tasneem and Alzubi, Saif and Zhang, Lei and Eldaly, Ahmed Karam and Ye, Xujiong},
  journal={arXiv preprint arXiv:2603.24388},
  year={2026}
}

@incollection{lopez2019medical,
  title={Medical image segmentation using deep learning},
  author={L{\'o}pez-Linares Rom{\'a}n, Karen and Garc{\'\i}a Oca{\~n}a, Mar{\'\i}a Inmaculada and Lete Urzelai, Nerea and Gonz{\'a}lez Ballester, Miguel {\'A}ngel and Mac{\'\i}a Oliver, Iv{\'a}n},
  booktitle={Deep learning in healthcare: paradigms and applications},
  pages={17--31},
  year={2019},
  publisher={Springer}
}

@article{veta2014breast,
  title={Breast cancer histopathology image analysis: A review},
  author={Veta, Mitko and Pluim, Josien PW and Van Diest, Paul J and Viergever, Max A},
  journal={IEEE transactions on biomedical engineering},
  volume={61},
  number={5},
  pages={1400--1411},
  year={2014},
  publisher={IEEE}
}

@article{tajbakhsh2020embracing,
  title={Embracing imperfect datasets: A review of deep learning solutions for medical image segmentation},
  author={Tajbakhsh, Nima and Jeyaseelan, Laura and Li, Qian and Chiang, Jeffrey N and Wu, Zhihao and Ding, Xiaowei},
  journal={Medical image analysis},
  volume={63},
  pages={101693},
  year={2020},
  publisher={Elsevier}
}

@article{kar2021review,
  title={A review on progress in semantic image segmentation and its application to medical images},
  author={Kar, Mithun Kumar and Nath, Malaya Kumar and Neog, Debanga Raj},
  journal={SN computer science},
  volume={2},
  number={5},
  pages={397},
  year={2021},
  publisher={Springer}
}

@inproceedings{long2015fully,
  title={Fully convolutional networks for semantic segmentation},
  author={Long, Jonathan and others},
  booktitle={Proceedings of the IEEE conference on computer vision and pattern recognition},
  pages={3431--3440},
  year={2015}
}

@inproceedings{ronneberger2015u,
  title={U-net: Convolutional networks for biomedical image segmentation},
  author={Ronneberger, Olaf and others},
  booktitle={International Conference on Medical image computing and computer-assisted intervention},
  pages={234--241},
  year={2015},
  organization={Springer}
}

@article{vaswani2017attention,
  title={Attention is all you need},
  author={Vaswani, Ashish and others},
  journal={Advances in neural information processing systems},
  volume={30},
  year={2017}
}

@article{dosovitskiy2020image,
  title={An image is worth 16x16 words: Transformers for image recognition at scale},
  author={Dosovitskiy, Alexey and Beyer, Lucas and Kolesnikov, Alexander and Weissenborn, Dirk and Zhai, Xiaohua and Unterthiner, Thomas and Dehghani, Mostafa and Minderer, Matthias and Heigold, Georg and Gelly, Sylvain and others},
  journal={arXiv preprint arXiv:2010.11929},
  year={2020}
}

@article{xiao2023transformers,
  title={Transformers in medical image segmentation: A review},
  author={Xiao, Hanguang and Li, Li and Liu, Qiyuan and Zhu, Xiuhong and Zhang, Qihang},
  journal={Biomedical Signal Processing and Control},
  volume={84},
  pages={104791},
  year={2023},
  publisher={Elsevier}
}

@inproceedings{zheng2021rethinking,
  title={Rethinking semantic segmentation from a sequence-to-sequence perspective with transformers},
  author={Zheng, Sixiao and Lu, Jiachen and Zhao, Hengshuang and Zhu, Xiatian and Luo, Zekun and Wang, Yabiao and Fu, Yanwei and Feng, Jianfeng and Xiang, Tao and Torr, Philip HS and others},
  booktitle={Proceedings of the IEEE/CVF conference on computer vision and pattern recognition},
  pages={6881--6890},
  year={2021}
}

@article{chen2022letcp,
  title={LETCP: a label-efficient transformer-based contrastive pre-training method for brain tumor segmentation},
  author={Chen, Shoucun and Zhang, Jing and Zhang, Tianchi},
  journal={Applied Sciences},
  volume={12},
  number={21},
  pages={11016},
  year={2022},
  publisher={MDPI}
}

@inproceedings{he2022masked,
  title={Masked autoencoders are scalable vision learners},
  author={He, Kaiming and Chen, Xinlei and Xie, Saining and Li, Yanghao and Doll{\'a}r, Piotr and Girshick, Ross},
  booktitle={Proceedings of the IEEE/CVF conference on computer vision and pattern recognition},
  pages={16000--16009},
  year={2022}
}

@article{thisanke2023semantic,
  title={Semantic segmentation using vision transformers: A survey},
  author={Thisanke, Hans and Deshan, Chamli and Chamith, Kavindu and Seneviratne, Sachith and Vidanaarachchi, Rajith and Herath, Damayanthi},
  journal={Engineering Applications of Artificial Intelligence},
  volume={126},
  pages={106669},
  year={2023},
  publisher={Elsevier}
}

@inproceedings{strudel2021segmenter,
  title={Segmenter: Transformer for semantic segmentation},
  author={Strudel, Robin and Garcia, Ricardo and Laptev, Ivan and Schmid, Cordelia},
  booktitle={Proceedings of the IEEE/CVF international conference on computer vision},
  pages={7262--7272},
  year={2021}
}

@article{ding2023large,
  title={A large-scale synthetic pathological dataset for deep learning-enabled segmentation of breast cancer},
  author={Ding, Kexin and Zhou, Mu and Wang, He and Gevaert, Olivier and Metaxas, Dimitris and Zhang, Shaoting},
  journal={Scientific Data},
  volume={10},
  number={1},
  pages={231},
  year={2023},
  publisher={Nature Publishing Group UK London}
}

@inproceedings{berroukham2023vision,
  title={Vision transformers: a review of architecture, applications, and future directions},
  author={Berroukham, Abdelhafid and Housni, Khalid and Lahraichi, Mohammed},
  booktitle={2023 7th IEEE congress on information science and Technology (CiSt)},
  pages={205--210},
  year={2023},
  organization={IEEE}
}

@article{liu2021self,
  title={Self-supervised learning: Generative or contrastive},
  author={Liu, Xiao and Zhang, Fanjin and Hou, Zhenyu and Mian, Li and Wang, Zhaoyu and Zhang, Jing and Tang, Jie},
  journal={IEEE transactions on knowledge and data engineering},
  volume={35},
  number={1},
  pages={857--876},
  year={2021},
  publisher={IEEE}
}

@article{xie2021segformer,
  title={SegFormer: Simple and efficient design for semantic segmentation with transformers},
  author={Xie, Enze and others},
  journal={Advances in neural information processing systems},
  volume={34},
  pages={12077--12090},
  year={2021}
}

@article{mahbod2021cryonuseg,
  title={CryoNuSeg: A dataset for nuclei instance segmentation of cryosectioned H\&E-stained histological images},
  author={Mahbod, Amirreza and others},
  journal={Computers in biology and medicine},
  volume={132},
  pages={104349},
  year={2021},
  publisher={Elsevier}
}

\end{document}